\documentclass[11pt,a4paper]{article}
\usepackage{times,latexsym}
\usepackage{url}
\usepackage[T1]{fontenc}

\usepackage[acceptedWithA]{tacl2021v1}

\usepackage{xspace,mfirstuc,tabulary}

\newif\iftaclinstructions
\taclinstructionsfalse 
\iftaclinstructions
\renewcommand{\confidential}{}
\renewcommand{\anonsubtext}{(No author info supplied here, for consistency with
TACL-submission anonymization requirements)}
\newcommand{\instr}
\fi

\iftaclpubformat 

\else

\fi

\usepackage{graphicx}
\usepackage{pifont}
\usepackage{amsmath}
\usepackage{amsfonts}
\usepackage{amssymb}
\usepackage{cleveref}
\usepackage{makecell}
\usepackage{markdown}

\title{TANQ: An open domain dataset of table answered questions}

\author{
  Mubashara Akhtar $^*$ $^\diamond$ $^\dagger$ $^\ddagger$
  \and 
  Chenxi Pang $^*$ $^\dagger$
  \\
  Andreea Marzoca$^\dagger$
  \and
  Yasemin Altun$^\dagger$
  \and
  Julian Martin Eisenschlos$^\dagger$ $^\mathsection$
  \ \\
  $^\diamond$King's College London, UK \& ETH Zurich, Switzerland
  \\
  \texttt{mubashara.akhtar@ai.ethz.ch}
  \\
  $^\dagger$Google DeepMind, Switzerland $^\mathsection$Universidad Nacional de Córdoba, Argentina
  \\
  \texttt{\{chenxipang,andreeam,altun,eisenjulian\}@google.com}
}

\date{}
\begin{document}
\maketitle

\def\thefootnote{*}\footnotetext{Equal contributions.}\def\thefootnote{\arabic{footnote}}
\def\thefootnote{$^\ddagger$}\footnotetext{Work done during Google DeepMind internship.}\def\thefootnote{\arabic{footnote}}

\begin{abstract}
Language models, potentially augmented with tool usage such as retrieval are becoming the go-to means of answering questions. Understanding and answering questions in real-world settings often requires retrieving information from different sources, processing and aggregating data to extract insights, and presenting complex findings in form of structured artifacts such as novel tables, charts, or infographics.
In this paper, we introduce TANQ,\footnote{\small Dataset's at \href{https://github.com/google-deepmind/tanq}{github.com/google-deepmind/tanq}} 
the first open domain question answering dataset where the answers require building tables from information across multiple sources. We release the full source attribution for every cell in the resulting table and benchmark state-of-the-art language models in open, oracle, and closed book setups. 
Our best-performing baseline, Gemini Flash reaches an overall F$1$ score of $60.7$, lagging behind human performance by $12.3$ points.
We analyse baselines' performance across different dataset attributes such as different skills required for this task, including multi-hop reasoning, math operations, and unit conversions. 
We further discuss common failures in model-generated answers, suggesting that TANQ is a complex task with many challenges ahead.
%

\end{abstract}

\section{Introduction}

Understanding and solving problems in real-world scenarios
often requires reasoning across multiple documents and data modalities.
This includes $(i)$ retrieving information from different sources, $(ii)$ processing and aggregating data to extract insights, and $(iii)$ presenting complex findings in a structured format, for example a table or infographics, to communicate them to readers. 

\begin{figure}[h!]
    \centering
    \resizebox{0.48\textwidth}{!}{%
    \includegraphics{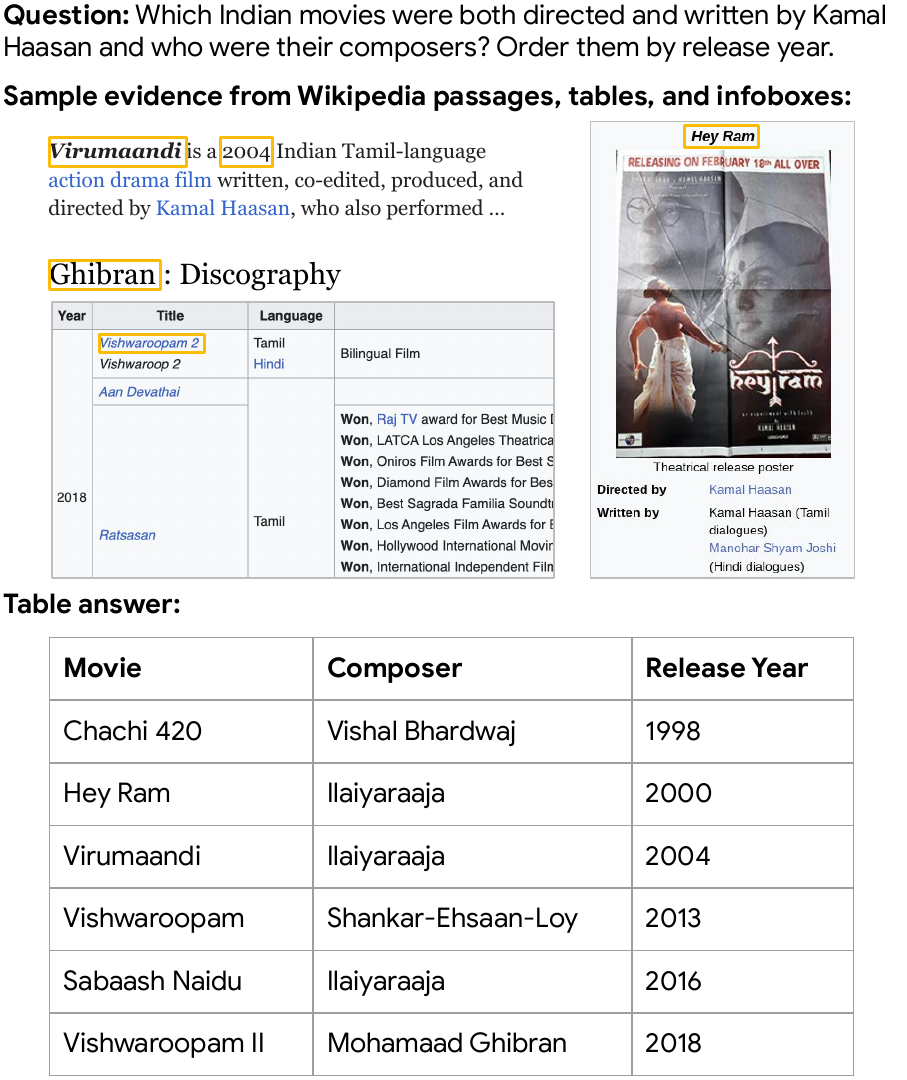}
    }
    \vspace{-1.5em}
    \caption{An example question in TANQ and its corresponding table answer. 
    Supporting evidence from multiple pages in a Wikipedia snapshot is provided for each data point inside the table.
    We highlight the rationale inside each snippet in yellow.
    LLMs are evaluated with or without access to the evidence.}
    \label{fig:tanq_example}
    \vspace{-1.5em}
\end{figure}

Previous studies show that knowledge workers across domains, e.g. finance, science, and economics, spend around $20\%$ of their time searching and gathering information from different files into one document to extract insights and consequently answer information-seeking questions~\citep{chui2012social}. 
As a result, one of most challenging tasks for workers is to aggregate data and turn it into insights.
Tables as a structured representation of data are ubiquitous in real-world sources and are commonly used to communicate complex information. Hence, they can be the perfect modality to answer complex questions.


\begin{table*}
        \centering
        \scalebox{0.8}{
        \begin{tabular}{l c c c c c c}
        \hline
        \bf Dataset	& \bf Open Domain & \bf Multi Doc & \bf Answer Type & \multicolumn{3}{c}{\bf Document Type}\\
        \hline
         & & & & \bf Text & \bf Table & \bf Infobox \\
        \hline
         InfoTabs~\citep{gupta-etal-2020-infotabs} & & & short text & & & \ding{51} \\
         FeTAQA~\citep{nan-etal-2022-fetaqa} & & & free form text & & \ding{51} & \\
         FinQA~\citep{chen-etal-2021-finqa} & & & short text & \ding{51} & \ding{51} & \\
         TATQA~\cite{zhu-etal-2021-tat} & & & short text & \ding{51} & \ding{51} & \\
         MultiHiertt~\cite{zhao-etal-2022-multihiertt} & &  & numeric & \ding{51} & \ding{51} & \\
         OTTQA~\citep{ChenCSWC21} & \ding{51} & & short text& & \ding{51} & \\
         NQ-TABLES~\citep{herzig-etal-2021-open} & \ding{51} & & short text& & \ding{51} & \\
         HybridQA~\cite{chen-etal-2020-hybridqa} & & \ding{51} & short text& \ding{51} & \ding{51} & \\
         MultiTabQA~\citep{pal-etal-2023-multitabqa} & & \ding{51} & table & & \ding{51} & \\
        \hline
         \bf TANQ & \ding{51} & \ding{51} & table & \ding{51} & \ding{51} & \ding{51} \\
        \hline
        \end{tabular}}
        \vspace{-0.5em}
        \caption{\small \label{tab:related_work} Comparison of TANQ to related (table) question-answering datasets.}
        \vspace{-1em}
\end{table*}

Large language models (LLMs), often enhanced with external tools, have become a primary method for various application such as answering questions.
State-of-the-art question-answering (QA) systems integrate LLMs in various ways, from decomposing complex queries to retrieving documents using external tools or generating context data from knowledge acquired during model training.
However, their evaluation is mostly limited to simple datasets, e.g. TabFact~\citep{ChenWCZWLZW20} or HotpotQA~\citep{yang-etal-2018-hotpotqa}, whose questions can be answered by reasoning over a single table or text document and generating a short text sequence as answer.
This limits the applicability of such systems to perform complex multi-step research explorations.
Moreover, it differs from real-world needs where relevant information can be spread across documents and represented in different forms (e.g. text or tables). More often than not, generating a short text as answer is not sufficient for complex information-seeking questions.
 
In this paper, we  investigate LLMs' capabilities in reasoning over multiple data sources and formats (i.e. text, tables, infoboxes) to answer entity-centric questions and generate table answers as structured artifacts.
To address these challenges, we introduce \textbf{TANQ}, an open-domain, multi-hop question-answering (QA) dataset.
TANQ requires retrieving and aggregating data from multiple documents to compile and communicate answers as tables.
To solve TANQ, models require different skills in addition to data retrieval such as filtering, maths, and name normalization. 
We create TANQ applying a five-step, automated data collection process. We use QAMPARI~\citep{amouyal-etal-2023} as seed dataset and Wikidata as well as the Wikipedia corpus as data sources. For automated evaluation of different data collection and processing substeps, we use PaLM-$2$~\citep{anil-etal-2023}.

We evaluate several state-of-the-art LLMs on TANQ, including close, oracle and open book evaluation settings.
Finally, we study model-generated answer tables and discuss common failure cases and challenges related to TANQ.
Our evaluation of models across skills can further inform future tools and evaluation setups for LLMs to improve models for complex, information-seeking questions. 

Our \textbf{contributions} are as follows: 

\noindent \textbf{(a)} We introduce TANQ, the first open-domain question-answering benchmark that requires building answers in form of tables from multiple information sources. 

\noindent \textbf{(b)} We benchmark state-of-the-art language models in oracle, open and closed book setups, reaching an overall F$1$ score of $60.7$ with our best-performing (oracle) baseline.

\noindent \textbf{(c)} We evaluate model performance across different dataset characteristics,
discuss challenges and common failure types.


\begin{figure*}
\centering
\includegraphics[scale=0.6]{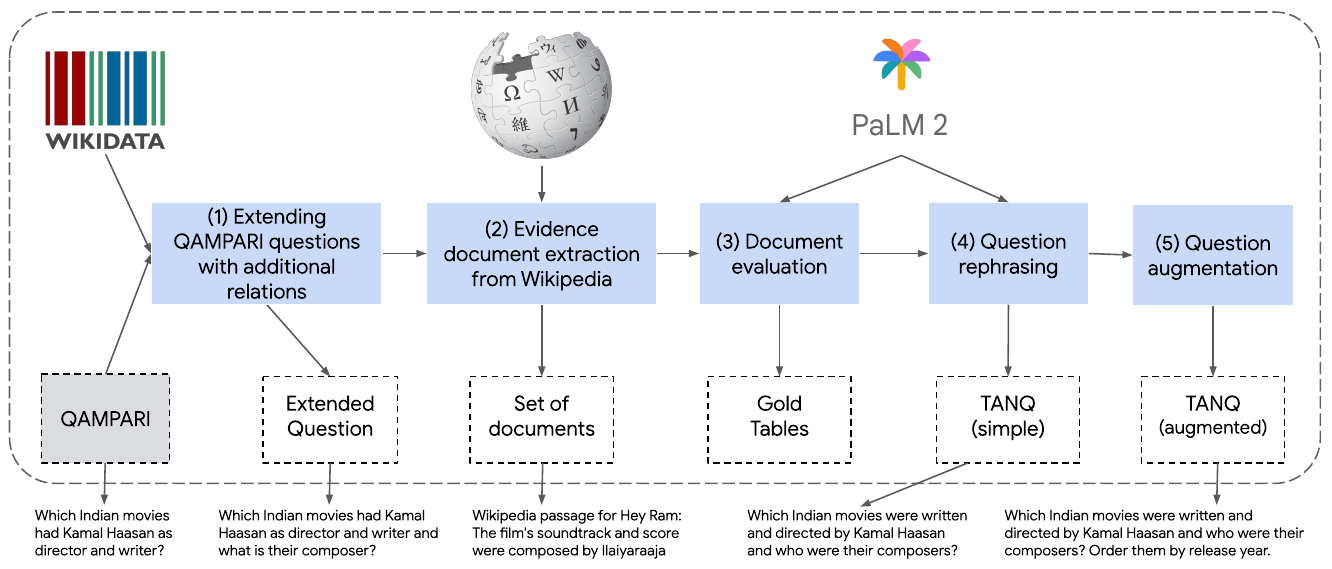}
\vspace{-0.7em}
\caption{\label{fig:tanq_pipeline} TANQ creation pipeline consisting of five steps: 1. Extending QAMPARI questions with additional relations based on Wikidata; 2. Evidence extraction (text, tables, infoboxes) from Wikipedia articles; 3. Evidence evaluation and (gold) answer table generation; 4. Rephrasing of question from first step; 5. Augmentation with additional skills to generate complex. We include a running example at the bottom.}
\vspace{-1.0em}
\end{figure*}

\section{Related Work}

Various benchmarks for QA have been released in recent years. Each one addresses different challenges related to the task. \Cref{tab:related_work} provides an overview and comparison of benchmarks. 

\paragraph{QA with text and/or table input.} 
A number of datasets use text and one or multiple tables for QA. While both text and tables have been considered as input modalities, the output of the datasets is mostly limited to short textual answers. 
HybridQA~\cite{chen-etal-2020-hybridqa}, for example, is a multi-hop QA dataset that requires reasoning over one table and multiple Wikipedia passages related to entities occurring in the table. HybridQA answers are short texts with location names being the most common answer types, followed by numbers, dates, and person names. 
Moreover, many QA benchmarks for reasoning over text and tabular context concentrate on the finance domain.
For example, the MultiHiertt~\cite{zhao-etal-2022-multihiertt} benchmark is created from financial reports. Questions require reasoning over texts and multiple tables. The answers are short numerical values with a focus on numerical reasoning. 
Other financial QA benchmarks are FinQA~\cite{chen-etal-2021-finqa} and TATQA~\cite{zhu-etal-2021-tat} where the context is one table and minimum two paragraphs related to the table. TATQA answers are short texts consisting of either one or multiple text spans from context paragraphs/tables or are free-form answers. 
%
MultimodalQA is a multi-hop, open-domain QA dataset that takes one table and related images and text paragraphs as input with answers similar to previously described datasets. 

\paragraph{Open-domain benchmarks.}
Most of the earlier described datasets have context provided in form of text and/or tables, whereas open-domain QA datasets first require extracting the relevant context, before answering the given question. 
The majority of open-domain QA datasets, such as WikiQA~\cite{yang-etal-2015-wikiqa}, TriviaQA~\cite{joshi-etal-2017-triviaqa}, RobustQA~\cite{han-etal-2023-robustqa}, are limited to textual context.
NQ Tables~\citep{herzig-etal-2021-open} extends table-QA to an open-domain setting where first top-$k$ tables are retrieved from a given corpus. These tables are processed by a reader component for generating the correct short-text answer. 
Built on HybridQA, the Open Table-and-Text Question Answering (OTT-QA) benchmark~\citep{ChenCSWC21} extends this setting by requiring to extract both tables and texts given multi-hop questions.

\paragraph{QA with table answers.}
The work closest to ours with respect to input and output modalities, is MultiTabQA~\cite{pal-etal-2023-multitabqa}. 
MultiTabQA seeds on the Spider dataset, a text-to-SQL dataset containing SQL queries, database tables, and natural language translations of the queries. 
\citet{pal-etal-2023-multitabqa} use table names occurring in SQL queries to extract the input tables and query Spider databases for answer table generation. 
While the dataset also generates answers as tables, it has certain limitations:
$(i)$ the benchmark input is limited to tables; $(ii)$ the input and output tables are highly structured database tables which differ from real-world scenarios where tables occur in documents and websites in various formats;
$(iii)$ the questions are limited to SQL-based queries.


\begin{table*}
        \centering
        \scalebox{0.8}{
        \begin{tabular}{l c}
        \hline
        \bf Question Type	& \bf Example Question\\
        \hline
         \textbf{1.} Simple & \makecell[l]{Which Belgian Grand Prix did Michael Schumacher win, and when did that happen?}\\
        \hline
         \textbf{2.} Intersection & \makecell[l]{For which movie did Chris Columbus receive credits as \textbf{both director and writer} and \\what was their composer, publication year, duration, and genre?   }\\
        \hline
         \textbf{3.} Composition & \makecell[l]{\textbf{Who choregraphed a work that was produced by} the Royal Ballet? What was their\\ date of birth, place of birth, occupation, and which awards did they receive?}\\
        \hline
        \bf Skill	& \bf  \\
        \hline
         \textbf{1.} No skill & \makecell[l]{What filmmaker directed a movie written by Val Guest and what is their place of birth, \\date of birth, occupation, date of death, and place of death?}\\
        \hline
        \textbf{2.} Filtering numeric & \makecell[l]{Which film was directed and produced by Mel Brooks and what was their composer \\ and duration? \bf Filter the answer table for duration equal to or larger than 88 minutes.} \\
        \hline
        \textbf{3.} Filtering time & \makecell[l]{Which Italian footballer transferred to Pro Sesto in summer of 2020 and what is their \\date of birth, place of birth? \bf Filter the answer table for date of birth equal to \\ \bf or after 1985.}\\
        \hline
        \textbf{4.} Filtering entity & \makecell[l]{What pieces of writing did Gregory Benford edit? What were their publication dates \\and publishers? \bf Filter the answer table for publisher equal to Bantam Books.}\\
        \hline
         \textbf{5.} Date-to-year conversion & \makecell[l]{Who were the members of Black Sabbath and what was their \textbf{year of birth}, genre,\\ instrument, and occupation? } \\
        \hline
         \textbf{6.} Quantity conversion & \makecell[l]{What work did Michael Mann write and direct and what was their publication date,\\ \textbf{duration in hours}, genre, director of photography?} \\
        \hline
         \textbf{7.} Time calculation & \makecell[l]{Which governor of Connecticut died while in office? What was their place of birth, \\occupation, date of death, date of birth, political party and \textbf{how many years did they live}?}\\
        \hline
         \textbf{8.} Approximation & \makecell[l]{What are the townships in Harper County, Kansas and what is their population\\ \textbf{rounded to the nearest ten}?}\\
        \hline

        \end{tabular}}
        \caption{\small \label{tab:tanq_skills_overview} An overview of TANQ question types and skills we use for augmenting the questions for more complex reasoning in the final step of the TANQ pipeline in \Cref{fig:tanq_pipeline}.}
        \vspace{-1.0em}
\end{table*}

\section{Building the TANQ benchmark}

TANQ evaluates the capability to answer open domain, multi-hop questions by aggregating data and generating answer tables.
%

\subsection{Task definition}

A TANQ dataset instance is a triple $(q, t, D)$, consisting of an entity-centric question $q$, a table answer $t$, and a document set $D$ (see \Cref{fig:tanq_example}).
To answer the multi-hop question $q$, first multiple \textit{sub-answers} are extracted from the document set $D$. 
The answer is generated in table form $t = \{t_{i, j}|i\le n, j\le m\}$ consisting of $n$ rows (i.e. one per extracted entities) and $m$ columns. 
The documents in $D$ 
provide supporting evidence for each cell of the answer table $t_{i, j}$. 
$D$ is either provided as input to models (oracle setting) or retrieved from the Wikipedia corpus as $D'$ (open book) and can consist of texts, tables and infoboxes.
%

\subsection{Preliminaries}

\textbf{QAMPARI.}
We use QAMPARI as seed dataset which is an open-domain QA dataset with lists of entities as answers~\citep{amouyal-etal-2023}. QAMPARI further includes Wikipedia text as supporting evidence for each entry of the answer list. 
Different to prior QA datasets with short textual answers, they align to natural questions which require a list of answers extracted from multiple sources. The dataset is semi-automatically created with Wikidata and Wikipedia as data sources and evaluated by human annotators.

\textbf{WikiData.}
Following \citet{amouyal-etal-2023}, we use Wikidata (WD) as a source for question generation. Wikidata~\citep{ErxlebenGKMV14} is a collaborative knowledge graph consisting of triples of entities and relations, i.e. $(e_1, r, e_2)$. Entities $e_i$ are values (e.g. $1990$) or items that represent a real-world concept, object, or a 
topic.\footnote{\href{https://en.wikipedia.org/wiki/Wikidata}{https://en.wikipedia.org/wiki/Wikidata}}
Relations are edges connecting entities, e.g. \texttt{(DonaldTrump, instanceOf, human)}. Whereas \texttt{Donald Trump} is the head entity and \texttt{human} the tail entity~\citep{krishna-etal-2022-proofver}.
We follow the notion for formal queries over Wikidata introduced by \citet{amouyal-etal-2023} for QAMPARI.
Applying a relation $r$ as query over WD items $e_i$ results in a set of (tail) entities: $[[r(e)]] = \{e_i | (e_i , r, e) \in\text{WD}\}$.

\textbf{Wikipedia.}
We extract supporting evidence in the form of sentences, tables and infoboxes (entity-tables at the top right corner of Wikipedia articles) from Wikipedia (WP).

\begin{table}
\centering
\scalebox{0.75}{
\begin{tabular}{l c c | l c c}
\hline  
\textbf{Type} & \textbf{\#} & \textbf{\%} & \textbf{Type} & \textbf{\#} & \textbf{\%}\\ \hline
WD Item & 57.6k & 79.1 & Numeric & 3.2k & 4.4 \\
Time & 11.9k & 16.4 & Text & 46 & 0.1 \\
\hline
\end{tabular}}
\caption{\label{tab:entities_types} Types of Wikidata entities we use in the $1$st pipeline step (see \Cref{fig:tanq_pipeline}) to extend QAMPARI and generate TANQ questions. We mostly use Wikidata (WD) items.}
\vspace{-1em}
\end{table}

\begin{table}
\centering
\scalebox{0.75}{
\begin{tabular}{l c c}
\hline  
\textbf{Question Type} & \textbf{Count} & \textbf{Freq (\%)}\\ \hline
1. Simple & 428 & 39.9 \\
2. Composition & 164 & 15.3 \\
3. Intersection & 482 & 44.9 \\
\hline
\bf Skill &  &  \\
\hline
1. No skill & 160 & 14.9 \\
2. Filtering numeric & 66 & 6.1 \\
3. Filtering time & 190 & 17.7 \\
4. Filtering entity & 268 & 25.0 \\
5. Date-to-year conversion & 230 & 21.4 \\
6. Quantity conversion & 64 & 6.0 \\
7. Time calculation & 23 & 2.1 \\
8. Approximation & 73 & 6.8 \\
\hline
\bf Skills per question &  &  \\
\hline
1. No skill & 300 & 27.9 \\
2. One skill & 657 & 61.2 \\
3. Two skills & 94 & 8.8 \\
4. Three skills & 23 & 2.1 \\
\hline
\end{tabular}}
\caption{\label{tab:skill_types} Question types and skills in TANQ. \textit{Simple} denotes questions which require neither composition nor intersection. See~\Cref{tab:tanq_skills_overview} for exemplary questions.}
\vspace{-1em}
\end{table}

\subsection{TANQ benchmark pipeline}

\Cref{fig:tanq_pipeline} provides an overview of the TANQ pipeline outlining all steps for data collection, processing and evaluation. 
We use QAMPARI, Wikidata and Wikipedia, as well as PaLM-2~\cite{anil-etal-2023}
for paraphrasing and validation.\footnote{Prompts provided in the appendix (\Cref{fig:evidence_eval_prompt} and~\ref{fig:rephrasing_prompt}).}
%

\paragraph{Step 1. Extending QAMPARI questions}
Starting with the QAMPARI questions and answer lists of entities, we extend each question $q$ with additional WD relations using the answer entities $e_a$. 
QAMPARI questions are classified in either simple, composition (e.g. \textit{``Who directed movies screen-written by Steven Spielberg?''}) or intersection (e.g. the example in \Cref{fig:tanq_example}).
We first query the WD knowledge graph to extract additional relations $r_{ext}$ linked to $e_a$, i.e. $r[e_a] = \{e_{ext} | (e_a, r_{ext}, e_{ext})\in\text{WD}\}$. 
Hence, each extension is a WD triple linking the QAMPARI answer $e_a$ (e.g. \textit{Hey Ram} in \Cref{fig:tanq_example}) through the relation $r_{ext}$ (i.e. \textit{composer})  to a new extended entity
$e_{ext}$, i.e. \textit{Ilaiyaraaja} for relation \textit{composer} and answer entity \textit{Hey Ram} in \Cref{fig:tanq_example}. 
We only select a relation which fulfils two conditions. 
First, it is part of a predefined relation set $R$, i.e. $r_{ext} \in R$.
We manually specify $R$ based on the WD relation used to create QAMPARI questions.
Second, the relation exists for all answers $e_a$ of question $q$.
Given $n$ extension relations, which fulfil these conditions, we extend the question $q$ in a template-style fashion: ``$[q]$ and what is their $[r_{ext1}], [r_{ext2}], [...]$ and $[r_{extn}]$.''
For example, the question in Fig. 1, was generated based on the initial question \textit{``Which Indian movies were both directed and written by Kamal Haasan?''} through extending with the WD relations \textit{composer} and \textit{release year}.
Additionally, we extract all extension entities $e_{ext}$ of the extension triple. 

\paragraph{Step 2. Evidence extraction from Wikipedia}
Next, we collect for each extension triple ($e_a$, $r_{ext}$, $e_{ext}$) supporting evidence using Wikipedia as an evidence source. 
Hereby, we search for supporting text, tables and infoboxes in the WP articles of $e_a$ and $e_{ext}$.
We apply simple heuristics and search for mentions of $e_a$ in the $e_{ext}$ article and vice versa. Moreover, we extend our queries with additional (heuristic-based) query words, for example considering different formats of how numbers and dates are represented in queries.

\paragraph{Step 3. Evidence evaluation \& answer table generation}
To evaluate the correctness of the previously collected evidence texts, tables and infoboxes, we employ PaLM-2 as an evidence evaluator. 
We prompt the LLM to evaluate the extracted evidence in a natural language inference setting.
For each extension triple $(e_a$, $r_{ext}$, $e_{ext})$, we construct template-based sentences $s$: ``$\langle e_a\rangle  \langle r_{ext}\rangle \langle e_{ext}\rangle$'', e.g. \textit{``Hey Ram composer Ilaiyaraaja''} for row two in \Cref{fig:tanq_example}.
We query the LLM to label the sentence s as ``supported'', ``refuted'', ``not enough information'' based on the provided evidence in form of a sentence, table or infobox entry. 
We then only consider the triples supported by at least one piece of evidence.\footnote{See \Cref{fig:evidence_eval_prompt} for exact prompt we used.}

\begin{figure*}[t]
    \centering
    \resizebox{0.83\textwidth}{!}{%
    \includegraphics{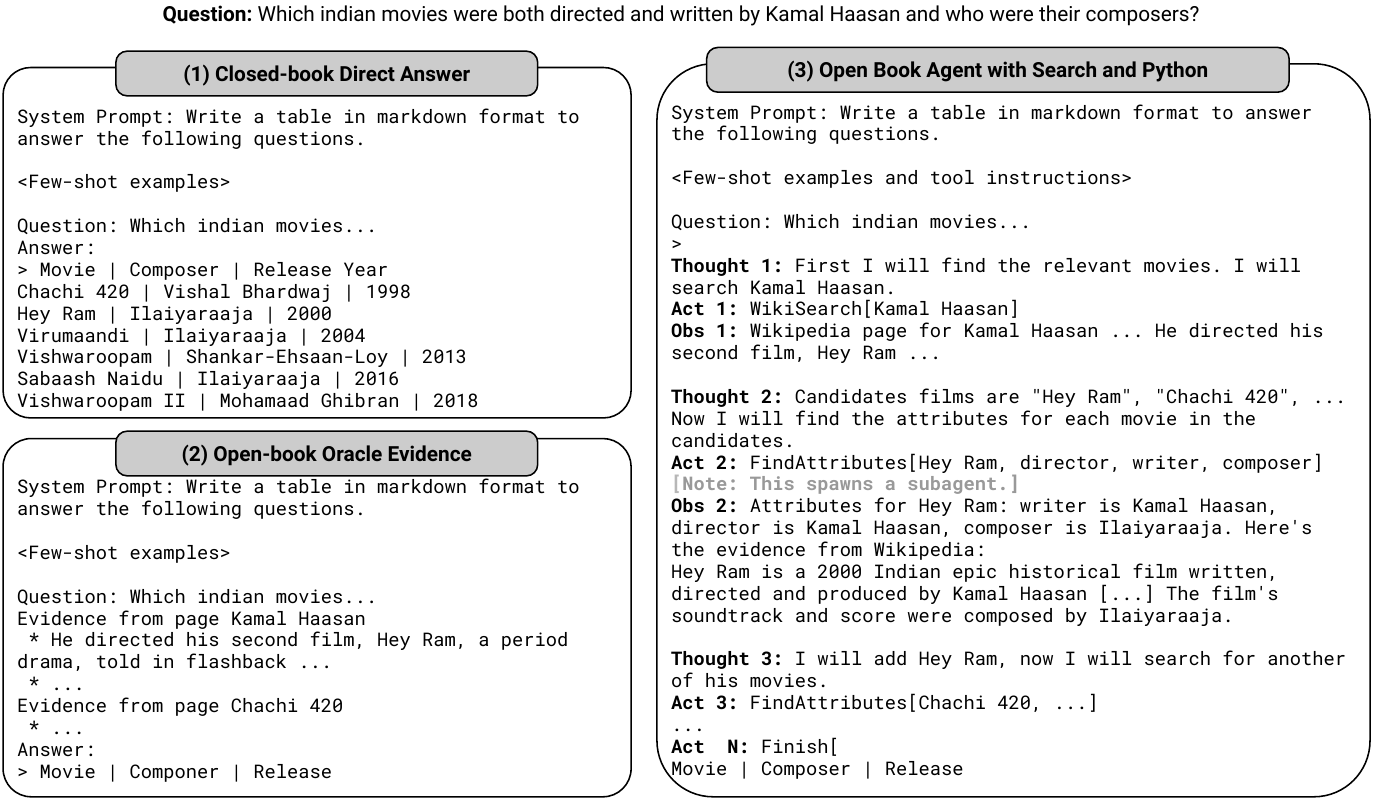}
    }
    \caption{Prompts for baseline evaluation on TANQ: closed book, oracle, and open book with augmented tools.}
    \label{fig:tanq_models}
    \vspace{-3mm}
\end{figure*}

\textbf{Answer table.} 
Each extended relation corresponds to a column in the answer table (e.g. column \textit{``Composer''} in \Cref{fig:tanq_example}). Since the question in \Cref{fig:tanq_example} is extended with two additional relations (i.e. \textit{composer} and \textit{release year}), the resulting answer table has three columns.
Hence, each cell in the answer table corresponds to a WD triple. For example, \texttt{(Hey Ram, composer, Ilaiyaraaja)} for cell \textit{``Ilaiyaraaja''}. 
%
Some generated answer tables contained multiple entries in a single cell, as seen in the \texttt{genre} column of \Cref{tab:metrics_example}. To generate realistic tables, we filtered out samples with more than five entries in any single cell, resulting in a test set of 1,074 TANQ samples for evaluation.

\paragraph{Step 4. Question rephrasing}
This step increases the naturalness of template-based extension questions generated in Step 1. 
Similarly to the earlier step, we prompt a PaLM-2 model in a few-shot setting. 
To ensure the question meaning is preserved during rephrasing, we add structured annotations in parenthesis to questions with the name of each relation (e.g \textit{``Which Indian movies were both directed (directed\_by) and written by (written\_by) Kamal Haasan?''}). 
We run up to 5 iterations of rephrasing and stop if all relations are present, or discard the question otherwise.\footnote{See \Cref{fig:rephrasing_prompt} for exact prompt we used.}

\paragraph{Step 5. Augmenting with skills}
Finally, to generate more challenging questions requiring further reasoning capabilities beyond retrieval, we extend TANQ questions by asking for realistic post-processing steps.
Overall, we augment questions with the following additional skills:
$(i)$ \textbf{Filtering} of answer table given a numeric, time, or attribute condition (rows $2.-4$ in \Cref{tab:tanq_skills_overview});
$(ii)$ \textbf{Conversion} of numbers, dates, locations and corresponding units (rows $5.-7$ in \Cref{tab:tanq_skills_overview});
$(iii)$ \textbf{Calculation} and introduction of an additional table column based on time attributes (e.g. \textit{``lifespan''} in the 8. row in \Cref{tab:tanq_skills_overview}). 
$(iv)$ \textbf{Approximating} numbers to the nearest ten, hundred, etc., see last row (\Cref{tab:tanq_skills_overview}).
We use up to three distinct skills for augmenting TANQ questions.
\Cref{tab:skill_types} provides a breakdown of TANQ dataset samples across skills. 



\subsection{Dataset statistics and analysis}

Overall, the generated TANQ dataset has $1395$ entries, $36.1\%$ of question type \textit{simple}, $40.9\%$ \textit{intersection}, and $22.9\%$ \textit{composition} questions.
We further plan to release a TANQ training set with approximately $42k$ samples. 
Approximately $72.4\%$ dataset instances (i.e. $1,010$) require at least one additional skill to answer the question. See~\Cref{tab:skill_types} for a breakdown of skills and question types in TANQ.
%
TANQ questions have an average a length of $21$ tokens and require extracting information about three relations on average. 
On average, TANQ answer tables have $6.7$ rows and four columns. See \Cref{tab:entities_types} for WD entity types used for extending QAMPARI questions.

\subsection{Manual evaluation of the pipeline}

We manually evaluated $100$ TANQ samples to assess if noise introduced through the dataset's automated generation pipeline. The focus of the analysis was twofold. First, examining the initial questions sourced from the QAMPARI dataset and, second, evaluating the generation of TANQ questions over the different pipeline steps. We identified five issues which are discussed below. 

\paragraph{Propagation of QAMPARI issues.}

While occurring only for a small subset of questions, one observation was that some errors present in the initial QAMPARI questions were propagated through the pipeline without being corrected. In $2$ out of the $100$ samples, the original QAMPARI questions contained issues, one was carried into TANQ while the other was fixed. Examples include:
\textit{``Robert Benton is the screenwriter and director of which software?''} This question confuses a film with software as we see by considering the requested attributes, i.e., screenwriter and director.

While these logical errors were carried forward into TANQ grammatical errors, for example, were fixed during rephrasing with LLMs: 
\textit{``Black Sabbath had who as a member?''} — issues phrasing that was later improved to \textit{``Who were the members of Black Sabbath [...]?''}.

\paragraph{Partially revealed answers.}

In three out of hundred evaluted questions, the question itself partially contained the answer, rendering the query less meaningful:
\textit{``Eon Productions and Harry Saltzman produced what series and what was their genre, director of photography, duration in second, producer, country of origin, composer, publication date, narrative location, screenwriter, and director?''} — in this case, the ``producer'' is already mentioned in the question.



\paragraph{Grammar and expression errors.}

Minor grammatical issues introduced through the TANQ pipeline, were found in $2$ out of the $100$ samples. In one cases, this was fixed during the rephrasing process, while other remained unresolved. For example,
\textit{``+89 minute''} was rephrased to \textit{``89 minutes''} in the sentence, \textit{``Which film was directed and produced by Mel Brooks and what was their composer, genre, duration? Filter the answer table for duration equal to or larger than +89 minute.''}
However, \textit{``Larger than 89 minutes''} remained, but \textit{``longer than 89 minutes''} would be more appropriate for this sentence.


\paragraph{Ordering of relations.}

While our evaluation metrics is not sensitive to the ordering of columns within the answer table, the order of relation appeared confusing for some questions. This occurred for seven out of the evaluated $100$ examples as shown below:
\textit{``What filmmaker directed a movie written by Val Guest and what was their date of birth, place of death, date of death, and place of birth?''} — the order of personal attributes (i.e., date and place of birth, date and place of death) seems unnatural as date of birth would naturally be followed by place of birth.

\paragraph{Ambiguous or unclear relations.}

In four out of the $100$ samples, certain relations were ambiguous or unclear, causing confusion about what exactly the question was asking for. Such as, \textit{``Who directed the movie which had the screenwriting done by Marc Norman and what was their date of birth, language spoken, written or signed, place of birth, sex or gender, country of citizenship, occupation?''} — here, the attribute \textit{``language spoken, written or signed''} is not clearly related to the director in question.



While we observed five different issues during the manual evaluation of the TANQ pipeline, the occurred only in a limited number of samples out of $100$ evaluated one. 
Thus, while these errors are present, they are not widespread enough to strongly impact the overall quality and utility of the TANQ dataset.

\begin{table*}
        \centering
        \scalebox{0.70}{
        \begin{tabular}{l c c c | c c c | c c c | c c c}
        \hline
        \bf Model	& \multicolumn{3}{c|}{\bf simple} & \multicolumn{3}{c|}{\bf composition} & \multicolumn{3}{c|}{\bf intersection} & \multicolumn{3}{c}{\bf all}\\
        \hline
         \bf  & \bf  P & \bf  R & \bf F1 & \bf  P & \bf  R & \bf F1 & \bf  P & \bf  R & \bf F1 & \bf  P & \bf  R & \bf F1  \\
        \hline
         \multicolumn{13}{c}{\bf Oracle setting} \\
        \hline
         Gemini Pro & 41.5 & 38.1 & 37.6 & 39.8 & 36.7 & 36.6 & 42.4 & 40.6 & 40.0 & 41.5 & 38.9 & 38.4 \\
         Gemini Flash & 66.4 & 59.0 & \textbf{60.4} & 67.4 & 58.4 & \textbf{60.4} & 66.8 & 59.3 & \textbf{61.1} & 66.8 & 59.0 & \textbf{60.7} \\
         GPT$4o$ & 58.2 & 47.0 & \underline{49.1} & 50.7 & 45.8 & \underline{44.5} & 53.1 & 42.9 & \underline{44.2} & 54.3 & 44.9 & \underline{45.9} \\
         Human & & & & & & & & & & 81.2 & 74.0 & \textbf{73.0} \\
        \hline
         \multicolumn{13}{c}{\bf Closed book setting} \\
        \hline
         PaLM-$2$ & 55.5 & 48.1 & \textbf{49.6} & 50.6 & 45.7 & \textbf{46.6} & 52.1 & 45.5 & \textbf{46.6} & 52.9 & 46.4 & \textbf{47.6} \\
         Gemma & 38.9 & 26.7 & 27.9 & 37.4 & 25.8 & 26.6 & 39.3 & 26.2 & 27.7 & 38.8 & 26.3 & 27.5 \\
         Gemini Pro & 46.9 & 30.3 & 31.1 & 43.8 & 29.5 & \underline{30.4} & 47.8 & 32.0 & \underline{33.5} & 46.6 & 30.9 & \underline{32.0} \\
         Gemini Flash &37.4 & 29.0 & \underline{31.3} & 36.2 & 27.2 & 29.6 & 36.6 & 28.8 & 30.7 & 36.8 & 28.5 & 30.7 \\
         GPT$4o$ & 46.8 & 26.6 & 29.1 & 28.1 & 16.0 & 18.1 & 35.9 & 22.3 & 24.4 & 37.9 & 22.4 & 24.6 \\
         \hline 
     \multicolumn{13}{c}{\bf Open book setting} \\
        \hline
         Tool LM & 24.2 & 20.3 & 18.4 & 43.6 & 33.0 & 36.8 & 50.8 & 44.2 & \textbf{46.6} & 39.7 & 33.4 & 34.5 \\
         \hline 
        \end{tabular}}
        \vspace{-0.5em}
        \caption{\small \label{tab:results_question_types} Baseline performance by question type. For all question types, we observe Gemini Flash ($60.7$ F$1$) and PaLM-$2$  ($47.6$ F$1$) to outperform other baselines in oracle and closed book setting respectively, lagging $12.3$ and $25.4$ points behind the human baseline of $73.0$.} 
\end{table*}

\begin{table*}
        \centering
        \scalebox{0.62}{
        \begin{tabular}{l c c c | c c c | c c c | c c c | c c c | c c c | c c c}
        \hline
        \bf Model	& \multicolumn{3}{c|}{\bf Filter num} & \multicolumn{3}{c|}{\bf Filter time} & \multicolumn{3}{c|}{\bf Filter entity} & \multicolumn{3}{c|}{\bf Date2Year} & \multicolumn{3}{c|}{\bf Quantity conv} & \multicolumn{3}{c|}{\bf Time calc} & \multicolumn{3}{c}{\bf Approx}\\
        \hline
         \bf  & \bf  P & \bf  R & \bf F1 & \bf  P & \bf  R & \bf F1 & \bf  P & \bf  R & \bf F1 & \bf  P & \bf  R & \bf F1 & \bf  P & \bf  R & \bf F1 & \bf  P & \bf  R & \bf F1 & \bf  P & \bf  R & \bf F1  \\
        \hline
         \multicolumn{21}{c}{\bf Oracle setting} \\
        \hline
         Gemini Pro & 32.8 & 28.9 & 29.0 & 40.7 & 38.6 & 38.1 & 42.7 & 39.9 & 39.0 & 39.5 & 36.2 & \underline{36.3} & 43.5 & 38.1 & 39.0 & 40.8 & 41.5 & 39.3 & 48.8 & 45.4 & \underline{45.2} \\
         Gemini Flash & 60.6 & 53.9 & \textbf{55.3} & 68.6 & 60.8 & \textbf{62.6} & 68.3 & 59.4 & \textbf{61.6} & 66.0 & 58.5 & \textbf{60.1} & 71.6 & 60.8 & \textbf{63.5} & 68.1 & 58.2 & \underline{59.6} & 69.3 & 65.2 & \textbf{65.5} \\
         GPT$4o$ & 56.0 & 47.0 & \underline{45.8} & 53.6 & 45.7 & \underline{47.7} & 55.9 & 46.0 & \underline{47.2} & 42.1 & 35.5 & 35.3 & 47.5 & 41.1 & \underline{43.2} & 81.6 & 65.8 & \textbf{71.8} & 34.2 & 35.3 & 34.3 \\
        \hline
         \multicolumn{21}{c}{\bf Closed book setting} \\
        \hline
         PaLM-$2$ & 50.5 & 46.3 & \textbf{46.9} & 53.9 & 47.1 & \textbf{48.0} & 54.9 & 48.5 & \textbf{49.7} & 51.4 & 44.7 & \textbf{46.1} & 51.5 & 45.2 & \textbf{46.8} & 53.1 & 48.3 & \textbf{50.1} & 52.3 & 43.4 & \textbf{45.7} \\
         Gemma & 37.4 & 26.5 & 27.8 & 37.1 & 26.4 & 27.3 & 40.3 & 26.2 & 27.6 & 40.4 & 27.4 & 28.7 & 37.4 & 23.8 & 25.2 & 39.2 & 27.4 & 29.3 & 40.8 & 27.5 & 29.6 \\
         Gemini Pro & 50.2 & 28.3 & 29.3 & 48.5 & 31.4 & \underline{32.4} & 46.3 & 31.2 & \underline{32.4} & 46.1 & 30.6 & \underline{31.9} & 48.0 & 31.4 & \underline{33.5} & 51.4 & 33.9 & \underline{35.3} & 42.8 & 30.2 & \underline{31.4} \\
         Gemini Flash & 38.7 & 29.5 & \underline{32.1} & 35.8 & 26.7 & 29.1 & 36.2 & 27.7 & 29.9 & 35.9 & 27.8 & 29.7 & 36.6 & 28.0 & 29.1 & 34.8 & 27.5 & 30.1 & 37.1 & 27.1 & 30.0 \\
         GPT$4o$ & 30.7 & 21.8 & 24.5 & 37.8 & 24.0 & 24.9 & 46.5 & 25.6 & 28.7 & 31.2 & 18.9 & 20.6 & 25.5 & 12.7 & 16.1 & 48.3 & 25.3 & 26.6 & 24.0 & 17.6 & 19.6 \\
         \hline 
         \multicolumn{21}{c}{\bf Open book setting} \\
         \hline 
         Tool LM & 48.7 & 39.1 & 43.1 & 36.1 & 29.7 & 30.3 & 37.6 & 33.0 & 34.2 & 53.1 & 37.8 & 41.9 & 33.8 & 28.4 & 30.2 & 25.8 & 18.2 & 21.1 & 48.3 & 44.0 & 45.9 \\
         \hline 
        \end{tabular}}
        \vspace{-0.5em}
        \caption{\small \label{tab:results_reasoning_skills} Baseline performance by skills required to answer the question: \textbf{Filter}ing with \textbf{num}erical/date\textbf{time}/\textbf{entity} conditions, date-to-year (\textbf{Date2Year}), quantity \textbf{conv}ersion, time \textbf{calc}ulation, and \textbf{approx}imation. In the oracle setting Gemini Flash performs best across all skills, while PaLM-$2$ performs better in the closed book evaluation.}
        \vspace{-1em}
\end{table*}

\section{Baselines \& Evaluation}

We evaluate TANQ in $(i)$ closed book, $(ii)$ oracle, and $(iii)$ open book setups. 
We give an overview of the different approaches in \Cref{fig:tanq_models}. 


\paragraph{Closed book.}
The closed book setup evaluates LLMs' capabilities in extracting relevant information acquired during training to answer TANQ questions.
For all experiments, we use the PaLM-$2$ Unicorn model~\citep{anil-etal-2023}, GPT$4o$~\citep{BrownMRSKDNSSAA20},
Gemini Pro and Flash~\citep{Gemini23}, as well as Gemma $9$B~\citep{Gemma2}.
We evaluate all baselines in a few-shot setting with three examples provided in the prompt (see top left prompt in \Cref{fig:tanq_models}). 

\paragraph{Oracle.}
In the oracle setup, we provide models source attribution for every cell of the answer table in form of oracle documents (i.e. text, tables, and infoboxes). Hence, the prompt consists of the TANQ question and multiple evidence sentences, infobox entries and/or tables (see bottom left prompt in \Cref{fig:tanq_models}). To provide further context, we include the Wikipedia page title, for text evidence also sub-/section titles, from where the evidence was extracted. Moreover, we add randomly selected oracle documents from other questions with similar attributes. This makes the oracle setting more challenging and requires models to filter the correct evidence from all provided ones first.
We exclude models with a context length of less than $4k$ from the oracle evaluation. The evidence samples require a longer context to be fully displayed. Otherwise, performance may decrease due to the input being cut off.




\paragraph{Open book.}
For the open book baseline, we extend the PaLM-$2$ Unicorn model with external tools for search and calculation. 
The Wikipedia-based search tools aim to mimic the human search approach on Wikipedia similar to \citet{YaoZYDSN023}. The agent model can access Wikipedia information through three tools: $(1)$
\texttt{WikiSearch(keywords):} a keyword based Wikipedia search that returns the Wikipedia article most relevant to the provided keywords; 
$(2)$ \texttt{FindEvidence(article, keywords):} a article-specific search that returns matched sentences, tables and infobox entries given keywords and a Wikipedia article; 
$(3)$ \texttt{GetIntro(article):} returns the introduction section of a Wikipedia article. 
For calculations, we provide the model a Python engine as an external tool, $(4)$ \texttt{Python(calculation)}.
We design the agent model to decompose and solve the TANQ task in multiple sub-tasks, consisting of $(i)$ retrieving requested entities in an iterative manner (e.g. movies in \Cref{fig:tanq_example}), $(ii)$ searching for entity-related information (e.g. release year), $(iii)$ post-processing information (e.g. filtering, calculation,, etc.), and $(iv)$ finally aggregating the information in form of a table. For some sub-tasks (e.g. entity retrieval), a separate sub-agent augmented with the required tools (e.g. \texttt{WikiSearch}) is spawned. 


\paragraph{Human baseline.}
We compare model performance against human performance based on $100$  answer tables generated by annotators. We provided the annotators the same input prompt as the models in the oracle setting (i.e. TANQ questions and evidence for each answer table cell). 

\paragraph{Evaluation metrics}
To evaluate answer tables, we adopted a version of the \emph{relative mapping similarity} (RMS) metrics introduced by~\citet{liu-etal-2023-deplot}. 
RMS views tables as unordered collections of mappings from row/colum headers to values. Hence, the metric is invariant to transpositions and column/row permutations. It allows small errors between tables keys/values of target and reference tables using the Normalized Levenstein Distance~\citep{BitenTMGRMJVK19}.
The metrics returns both precision and recall scores.

To evaluate the generated answer tables, we first converted table entries into a list of triplets. Each triplet consists of $(i)$ the entity name (given in the first column of the answer table), $(ii)$ the relation names (i.e., column names), and $(iii)$ the content of a table cell. If a table cell contains multiple values, such as in column \texttt{Genre} in \Cref{tab:metrics_example}, the cell content is split into multiple triplets with the same entity name and attribute name, but with different values extracted from the cell. For example, given \Cref{tab:metrics_example}, the resulting splitted triplets contains multiple triplets for genre:
\texttt{\{(Evita, duration, 129 min), (Evita, genre, biographical film)
(Evita, genre, musical), \dots\}}.

\begin{table}[h!]
\centering
\scalebox{0.7}{
\begin{tabular}{|l|l|l|l|}
\hline
\textbf{Film} & \textbf{Duration} & \textbf{Genre} & \textbf{Publication} \\ \hline
Evita & 129 min & \makecell[l]{biographical film,\\ musical} & 1996 \\ \hline
\end{tabular}}
\caption{\label{tab:metrics_example}Example TANQ answer table with multiple entries in a single cell, i.e., ``biological film, musical''.}
\vspace{-2mm}
\end{table}

After generating triplet lists for the gold table and the model-generated answer table, we calculate the similarity between them using relative distance for numbers and Normalized Levenshtein distance for text across each field in the triplet. 
Each triplet in the target is matched with its closest triplet in the prediction greedily and thus we can compute weighted precision and recall, in the same manner as~\citet{liu-etal-2023-deplot}.
The evaluation code will be released.

\section{Results \& Discussion}
\label{sec:results_discussion}

In this section, we address key research questions: $(1)$ Is TANQ a challenging dataset for state-of-the-art models? $(2)$ How do models perform in a closed book setting compared to using external context (oracle)? $(3)$ How effective are tool-augmented models on TANQ? $(4)$ What challenges arise from different TANQ specifications, e.g., question types, reasoning skills, etc.? (5) What are the common failure cases of the evaluated models?



\subsection{Question types}

In \Cref{tab:results_question_types}, we compare the performance of all oracle, close, and open book baselines.
In the oracle setting, we find Gemini Flash consistently outperforming other models with an overall F$1$ score of $60.7$, followed by GPT$4o$. However, still, a considerable gap remains to the \textbf{human baseline} of $73.0$. 
%
Both models experience a significant drop in performance in the closed book setting, falling behind the smaller, best-performing PaLM-$2$ model. Notably, PaLM-$2$ achieves higher recall scores, indicating that while the other models can generate some rows of the answer tables, they are slightly less accurate than PaLM-$2$, and their resulting tables are shorter, containing fewer entities.
Different to Gemini Flash/Pro and Gemma, GPT$4o$ and PaLM-$2$ struggle with more complex questions types, i.e., composition and intersection questions. 
Moreover, the $9$B-sized Gemma model demonstrates performance comparable to much larger models in closed book evaluation. 

\subsection{Evaluation with diverse prompts}
\label{sec:prompt_eval}

In addition to the evaluations discussed in \Cref{sec:results_discussion}, we further assessed the baselines to examine the impact of $(i)$ instruction tuning and $(ii)$ demonstration selection on the models' performance on TANQ. \Cref{tab:results_of_models_with_different_prompt_settings} compares the performance of baselines across different numbers of demonstrations in the input prompt (i.e., one-, three-, and five-shot), while \Cref{tab:results_of_Gemini_Flash_with_different_prompt_instruction_styles} provides an overview of how different instruction styles affect the performance of Gemini Flash.

For almost all baselines, performance improves slightly when the number of input examples is increased from one to three in the prompt. However, no further improvements are observed with an increase to five examples, supporting our decision to evaluate models in a $3$-shot setting.
Except for Gemini Pro and GPT$4o$, manually selecting input examples does not yield enhanced performance.

When comparing the performance of the top-performing oracle baseline, Gemini Flash, across different instruction styles, we find that the detailed (B), step-by-step (C), and simple (D) instruction styles yield similar performance (see \Cref{fig:rephrasing_prompt}). In contrast, using an empty instruction (A) - where we provide only input examples without additional details - results naturally in decreased performance (see \Cref{tab:results_of_Gemini_Flash_with_different_prompt_instruction_styles}). 
For our evaluation in \Cref{sec:results_discussion}, we used the simple instruction style, as additional details as per instruction B and C did not contribute to observable performance improvements on the evaluated models.

\subsection{Reasoning skills}
\begin{table}
        \centering
        \scalebox{0.62}{
        \begin{tabular}{l c c c | c c c | c c c }
        \hline
        \bf Model	& \multicolumn{3}{c|}{\bf One skill} & \multicolumn{3}{c|}{\bf Two skills} & \multicolumn{3}{c}{\bf Three skills} \\
        \hline
         \bf  & \bf  P & \bf  R & \bf F1 & \bf  P & \bf  R & \bf F1 & \bf  P & \bf  R & \bf F1  \\
        \hline
         \multicolumn{10}{c}{\bf Oracle setting} \\
        \hline
         Gemini Pro & 42.3 & 39.9 & 39.3 & 36.3 & 32.4 & 32.4 & 46.5 & 41.0 & 41.3 \\
         Gemini Flash &  66.8 & 58.6 & \textbf{60.6} & 67.4 & 59.4 & \textbf{60.8} & 74.5 & 68.8 & \textbf{70.1} \\
         GPT$4o$ & 51.5 & 42.4 & \underline{43.3} & 50.7 & 46.4 & \underline{47.6} & 51.0 & 39.7 & \underline{44.7} \\
        \hline
         \multicolumn{10}{c}{\bf Closed book setting} \\
        \hline
         PaLM-$2$ & 52.8 & 46.2 & \textbf{47.4} & 52.0 & 46.9 & \textbf{47.9} & 57.0 & 47.1 & \textbf{50.5} \\
         Gemma & 37.9 & 25.8 & 26.9 & 42.4 & 28.9 & 30.3 & 42.5 & 26.2 & 29.5 \\
         Gemini Pro & 46.9 & 30.4 & \underline{31.5} & 46.4 & 32.8 & \underline{34.4} & 50.1 & 30.2 & \underline{31.0} \\
         Gemini Flash & 35.6 & 27.8 & 29.9 & 37.4 & 28.3 & 30.6 & 39.5 & 23.2 & 25.9 \\
         GPT$4o$ & 39.4 & 24.7 & 26.5 & 32.5 & 14.5 & 17.6 & 9.4 & 1.8 & 3.0 \\
         \hline 
     \multicolumn{10}{c}{\bf Open book setting} \\
        \hline
         Tool LM & 40.7 & 33.5 & 35.0 & 46.4 & 35.6 & 39.8 & - & - & - \\
         \hline 
        \end{tabular}}
        \vspace{-0.5em}
        \caption{\small \label{tab:results_skill_count} Baseline performance by number of skills required to successfully answer the TANQ questions.}
        \vspace{-1em}
\end{table}

\Cref{tab:results_reasoning_skills} presents the performance of models on more challenging questions requiring further skills such as filtering and time conversion.
%




\begin{table*}
        \centering
        \scalebox{0.7}{
        \begin{tabular}{l c c c | c c c | c c c | c c c}
        \hline
        \bf Model	& \multicolumn{3}{c|}{\bf One Relation} & \multicolumn{3}{c|}{\bf Three Relations} & \multicolumn{3}{c|}{\bf Five Relations} & \multicolumn{3}{c}{\bf Ten Relations}\\
        \hline
         \bf  & \bf  P & \bf  R & \bf F1 & \bf  P & \bf  R & \bf F1 & \bf  P & \bf  R & \bf F1 & \bf  P & \bf  R & \bf F1  \\
        \hline
         \multicolumn{13}{c}{\bf Oracle setting} \\
        \hline
         Gemini Pro & 41.6 & 39.2 & 38.6 & 42.2 & 40.7 & 39.5 & 38.5 & 36.2 & 35.0 & 35.8 & 36.0 & \underline{35.8} \\
         Gemini Flash & 67.0 & 59.5 & \textbf{61.1} & 68.5 & 59.2 & \textbf{61.8} & 64.7 & 58.6 & \textbf{59.5} & 64.7 & 58.4 & \textbf{60.5} \\
         GPT$4o$ & 55.1 & 44.7 & \underline{45.8} & 63.2 & 53.6 & \underline{53.5} & 74.9 & 43.3 & \underline{50.2} & 23.8 & 19.4 & 20.5 \\
        \hline
         \multicolumn{13}{c}{\bf Closed book setting} \\
        \hline
         PaLM-$2$ & 53.7 & 47.4 & \textbf{48.4} & 51.8 & 45.2 & \textbf{46.0} & 54.2 & 47.7 & \textbf{48.8} & 62.4 & 47.7 & \textbf{48.9} \\
         Gemma & 38.5 & 26.3 & 27.6 & 35.6 & 25.6 & 26.5 & 47.6 & 30.5 & \underline{31.1} & 42.0 & 35.8 & 37.8 \\
         Gemini Pro & 45.6 & 32.6 & \underline{33.5} & 47.7 & 30.8 & \underline{31.4} & 51.1 & 29.4 & 29.9 & 43.9 & 39.2 & \underline{40.9} \\
         Gemini Flash & 36.6 & 28.6 & 30.7 & 37.7 & 28.4 & 31.0 & 35.3 & 26.9 & 28.6 & 36.1 & 26.1 & 29.4 \\
         GPT$4o$ & 37.5 & 21.9 & 24.4 & 33.6 & 25.1 & 27.8 & 55.6 & 15.4 & 18.2 & 34.3 & 21.3 & 23.2 \\
         \hline 
     \multicolumn{13}{c}{\bf Open book setting} \\
        \hline
         Tool LM & 31.6 & 32.0 & 30.2 & 42.4 & 33.7 & 36.7 & 34.0 & 32.9 & 33.0 & 18.4 & 17.8 & 18.1 \\
         \hline 
        \end{tabular}}
        \vspace{-0.5em}
        \caption{\small \label{tab:results_question_properties} 
        Baseline performance by number of relations in TANQ questions (i.e., corresponds to number of columns in the answer table). Most baselines demonstrate a stable performance as relations increase.}
        \vspace{-1em}
\end{table*}

\paragraph{Gemini.}
Overall, questions requiring filtering with numerical conditions poses the biggest challenge for Gemini models resulting in a performance drop compared to other skills. 
For example, Gemini Flash's performance drops from $60.7$ F$1$ (overall) to $55.3$ in oracle evaluation.

\paragraph{PaLM-$2$.} 
For PaLM-$2$, we observe that the model struggles particularly with questions 
requiring numeracy, i.e. approximation of numerical values, quantity conversion, calculations based on datetime attributes, and filtering with numerical conditions.
This challenge does not persist in the open book baseline where PaLM-$2$ is augmented with a calculator tool. 
Despite these limitations, the PaLM-$2$ model outperforms all other baselines across all skills in the closed book setting. 


\paragraph{GPT$4o$.}
For GPT-$4o$, no clear patterns are observed regarding its limitations in specific reasoning skills. In both the oracle and closed book scenarios, the model struggles with date-to-year conversions and the approximation of numerical values. However, it outperforms other closed book baselines in time calculations.

\paragraph{Number of skills.}
\Cref{tab:results_skill_count} shows baseline performance by the number of skills required to generate the correct answer table. 
We observe across baselines a stable performance as the number of skills increases and questions become more complex. Except GPT$4o$, which shows significant performance decreases as the number of skills increases in the close book setting, e.g., from $26.5$ (one skill) to $17.6$ (two skills).

\begin{table}[!h]
        \centering
        \scalebox{0.62}{
        \begin{tabular}{l c c c | c c c | c c c }
        \hline
        \bf Model	& \multicolumn{3}{c|}{\bf Short} & \multicolumn{3}{c|}{\bf Medium} & \multicolumn{3}{c}{\bf Long} \\
        \hline
         \bf  & \bf  P & \bf  R & \bf F1 & \bf  P & \bf  R & \bf F1 & \bf  P & \bf  R & \bf F1  \\
        \hline
         \multicolumn{10}{c}{\bf Oracle setting} \\
        \hline
         Gemini Pro & 47.0 & 41.2 & \underline{39.9} & 42.3 & 38.1 & 37.8 & 41.1 & 39.3 & 38.8 \\
         Gemini Flash & 68.6 & 56.6 & \textbf{58.3} & 66.2 & 56.0 & \textbf{58.0} & 67.2 & 60.6 & \textbf{62.2} \\
         GPT$4o$ & 62.9 & 31.6 & 36.3 & 53.1 & 37.4 & \underline{39.5} & 55.0 & 49.7 & \underline{50.0} \\
        \hline
         \multicolumn{10}{c}{\bf Closed book setting} \\
        \hline
         PaLM-$2$ & 56.4 & 43.9 & \textbf{43.7} & 54.5 & 45.0 & \textbf{45.9} & 51.9 & 47.3 & \textbf{48.5} \\
         Gemma & 35.8 & 25.8 & 25.6 & 37.6 & 25.1 & 25.9 & 39.4 & 27.0 & 28.4 \\
         Gemini Pro & 47.8 & 29.6 & \underline{29.7} & 45.9 & 31.5 & \underline{32.7} & 47.0 & 30.5 & \underline{31.6} \\
         Gemini Flash & 32.9 & 25.7 & 25.6 & 36.3 & 27.9 & 29.4 & 37.1 & 28.9 & 31.4 \\
         GPT$4o$ & 30.6 & 13.0 & 14.9 & 35.3 & 17.8 & 20.2 & 39.6 & 25.4 & 27.5 \\
         \hline 
     \multicolumn{10}{c}{\bf Open book setting} \\
        \hline
         Tool LM & 23.2 & 29.2 & 23.0 & 33.3 & 31.7 & 30.9 & 43.3 & 34.3 & 36.5 \\
         \hline 
        \end{tabular}}
        \vspace{-0.5em}
        \caption{\small \label{tab:results_table_len} Baseline performance by length of answer tables. \textbf{Short} tables have $< 3$ rows, \textbf{medium} table up to six, and \textbf{long} table $>7$ rows. Most baselines show little performance variation across table sizes.}
        \vspace{-1.5em}
\end{table}

\subsection{Relations in TANQ questions}
\Cref{tab:results_question_properties} demonstrates baselines' performance across the number of relations in the given question. For example, in \Cref{fig:tanq_example} two relations, i.e. composer and release year, are requested for the movies specified in the question. 
While multiple models show a performance decrease with an increasing number of relations, this gap is particularly significant for the tool augmented model where the F$1$ score decrease by almost twelve points comparing questions requiring one relations vs. ten. 
Moreover, we observe for most baselines (oracle/close/open book) little performance drops comparing questions with one relation to those with three - indicating that up to a certain limit models can successfully retrieve information for an increasing number of relations. However, the performance gaps increase for GPT$4o$ (close) and the tool-augmented model as the number of relations increases further, i.e. from three to five and from five to ten.

\subsection{Answer table length}
\Cref{tab:results_table_len} gives an overview of model performance across different answer table lengths: short answer tables (up to three rows), medium (up to six), and long tables (seven or more rows).
Models that perform well on longer tables are GPT-$4o$, Gemini Flash, and PaLM-$2$. As table size increases, so does the number of rows the models can correctly retrieve, resulting in higher recall scores, while precision slightly drops for most oracle baselines models.
%
It is apparent that the model struggles with extracting correct information from the growing list of provided oracle documents as the table size increases.
This is obviously not the case for close book evaluation as no oracle documents are given as input.   

\subsection{Failures types}

\begin{table}
\centering
\scalebox{0.73}{
\begin{tabular}{l l l l l}
\hline  
\textbf{Failure Type} & \bf Gemini & \bf PaLM-2 & \bf GPT$4o$  & \bf Tool LM\\ \hline
Rel missing & 37.3 & 33.3 & 0 & 3.9 \\
No header & 27.5 & 3.9 & 0 & 5.9\\
Filter issues & 15.7 & 15.8 & 4.0 & 5.9\\
Halluc. relations & 5.9 & 9.8 & 5.9 & 2.0 \\
Halluc. other & 0 & 15.7 & 0 & 7.8\\
Missing entity & 80.4 & 58.8 & 27.5 & 78.4\\
Non-table & 31.4 & 21.6 & 0 & 15.7\\
Partial answers & 3.9 & 23.5 & 2.0 & 0\\
Wrong answer & 14.9 & 7.8 & 9.8 & 13.7\\
\hline
\end{tabular}}
\vspace{-0.5em}
\caption{\label{tab:failures} Common failure types: \textbf{rel}ations \textbf{missing} in answer table, \textbf{no} column \textbf{header}, \textbf{filter} issues (e.g. filter condition ignored), \textbf{halluc}inated \textbf{relations}, \textbf{other} types of \textbf{halluc}inations, \textbf{missing entity} (i.e. row), output \textbf{not} in \textbf{table} format, \textbf{partial answers} in one or multiple table cells, \textbf{wrong answer} given in one or multiple cells. Scores given as $\%$ of all annotated samples.}
\vspace{-1.5em}
\end{table}
Hence, we further study common failures cases of oracle baselines. The aim is to understand challenges related to TANQ when the model has access to necessary information, in form of oracle documents.

We manually evaluate a subset of model predictions to identify common failure cases, categorized as follows: $1.)$ \textbf{Relation missing}: At least one expected relation (column) is missing; $2.)$ \textbf{No table header} generated; $3.)$ \textbf{Filtering}: Conditions (attribute, numeric, dates) ignored or incorrectly applied; $4.)$ \textbf{Hallucinated relations}: Unrequested columns are added; $5.)$ \textbf{Other hallucinations}; $6.)$ \textbf{Missing entity}: Expected rows are missing; $7.)$ \textbf{Non-table answers}: Text or non-table format returned; $8.)$ \textbf{Partial answers}: Incomplete information for a given entity or attribute; $9.)$ \textbf{Wrong answer (cell)}: Incorrect cell content.

Comparing open and oracle baselines, our observations show that the best performing model, w.r.t least failures, is GPT$4o$ (oracle). The only significant failure category we identify for GPT$4o$ are missing entities in generated answer tables. 
The most present issue we observe for Gemini Pro (oracle) are missing entities, missing relations, resulting in answer tables with a subset of columns, non-table outputs, and missing headers of generated tables. 
Similarly, multiple aspects pose a challenge for PaLM-$2$ (oracle), including hallucinations, missing entities/relations, and applying filtering conditions correctly. 
For the tool augmented model missing entities is a frequent issue.

\subsection{Detailed evaluation of open book model}

It is noteworthy that the open book, ReAct-based baseline performed worse than other baselines, also the closed book ones. To better understand the limitations of the open book baseline, we manually evaluated the reasoning chains generated by the model for $50$ TANQ samples. These chains outline the verbal reasoning processes and actions taken by the model in an interleaved manner.
Our evaluation identified six different areas where the open book model struggles.

\paragraph{Retrieval of a complete entity list}

The most prevalent issue, found in $19$ out of $50$ samples, was that the answer list did not contain all entities of the target table. Our review of the ReAct chain logs showed that the model often terminated the search for further entities after finding a list of entities. For example, it might start searching in the introduction section of a Wikipedia article and identify some movies directed by a particular director, but then fail to continue searching the rest of the page, missing additional films.

\paragraph{Generation of correctly-formatted answer tables.}

We also found issues related to answer table formatting in seven out of fifty samples. For instance, in one case, the name column was incorrectly replicated into a table row (i.e., ``Building'' below) in \Cref{tab:react_example_2}.

\begin{table}[h!]
\centering
\scalebox{0.74}{
\begin{tabular}{|l|l|l|}
\hline
 & name & located in \\ \hline
building & Old Main at the University of Arkansas & Arkansas \\ \hline
\end{tabular}}
\vspace{-0.6em}
\caption{\label{tab:react_example_2}Example issue related to formatting answer tables: The first column is redundant.}
\vspace{-1.2em}
\end{table}

\paragraph{Retrieval of attributes.}

In six out of fifty samples, the model retrieved entities but failed to find attributes for all entities, resulting in partly empty rows (see \Cref{tab:react_example_3}).

\paragraph{Answering intersection questions.}

A similar issue arises with intersection questions, which ask for entities that fulfill multiple conditions. For example, ``What science fiction films were both written and directed by Steven Spielberg? What is their genre, publication date, and duration?'' The open book model often produced tables that fulfilled only one of these conditions, such as listing films written by Steven Spielberg without considering if he also directed them. This problem occurred in five out of the fifty evaluated examples.

\paragraph{Answering composition questions.}

As the name suggests, composition questions require the model to first retrieve a list of intermediate entities, which are then used to compile the final entity list needed for the answer table. For example, consider the question: \textit{``Who directed the film that Jules Feiffer wrote? What is their occupation, date of death, date of birth, country of citizenship, place of birth, and award received?''} The model should first retrieve the films as intermediate entities to identify the requested directors. However, in four out of the fifty evaluated samples, the model generated answer tables that included intermediate entities only instead of the requested entities. For instance, the answer table for the above question looked as demonstrated in \Cref{tab:react_example_1}. The first row correctly shows the directors, but the second and third rows contain movies written by Jules Feiffer rather than their directors.

\paragraph{Applying filters to the generated table}

Finally, the model commonly ignored filtering conditions specified in the input question. We observed this in ten out of fifty samples. For example, in the question: \textit{``Who was a mayor of Saint Paul, MN, and what was their occupation, position held, and date of birth? Filter the answer table for date of birth equal to or before 1829,''} the resulting table included mayors born after $1829$, failing to apply the requested filter.

\section{Future Work}


Our results on TANQ highlight the need for qualitative benchmarks, metrics, and modeling approaches to address current limitations.

We initially used the RMS metrics from \citet{liu-etal-2023-deplot} to evaluate table generation. However, RMS was developed for the chart-to-table conversion task, which involves mostly numerical tables. In contrast, TANQ tables contain significant text content, leading to variations such as expressing ``USA'' as ``United States.'' 
Another difference is the more diverse table structure, where cells often contain lists of values. 
We adapted the RMS metrics for TANQ (see \Cref{sec:results_discussion}). 
Our experiments indicate that this version of RMS strikes a right balance of providing signal, being simple to explain and implement, and fast to execute. 
However, further research on table evaluation is necessary to consider a broader range of formats.

Interestingly, we found that smaller models like Gemma ($9$B parameters) performed surprisingly well. While Gemma was not among the top models for any specific skill or question type in the closed book setting, it did not fall far behind models that are significantly larger, such as PaLM-$2$, Gemini Pro, and GPT-$4o$. This suggests that, despite its smaller size, models like Gemma and Gemini Flash offer promising results, though there is still room for improvement. While the size of Gemini Flash is undisclosed, it is known to be smaller than Gemini Pro, further highlighting that scale alone does not guarantee better performance on complex reasoning tasks.This raises an important question for future research: how can we further enhance the performance of smaller models?

Additionally, while tool-based LMs are popular, our results suggest that they may not be the optimal choice in domains where up-to-date knowledge is not required, especially for answering simple questions. For more complex questions involving intersecting information, tool-augmented LMs perform comparably to other baselines, but for simpler questions, they lag behind both oracle and closed book baselines.

Numeracy remains a challenge for state-of-the-art models. Despite advancements in numerical reasoning with LLMs, many models still struggle with questions requiring simple numerical skills such as filtering the content of a based on a numerical condition. While progress has been made in recent years~\citep{imani-etal-2023-mathprompter, akhtar-etal-2023-exploring, chen-etal-2022-convfinqa}, challenges remain.

Another challenge is that all evaluated models tend to generate incomplete answer tables, often missing some entities (rows) and relations (columns) from the target table. Future work should focus on developing methods to address these limitations and generate answer tables that capture the entire requested information.
Addressing these issues and improving table evaluation metrics are important areas for future research.

\section{Conclusion}

This paper introduces TANQ, the first open domain question answering dataset where the answers require building tables from information across multiple sources.  To create TANQ, we design and apply an automated dataset pipeline using large language models and Wikipedia and Wikidata as knowledge sources. We further release for each cell of the answer tables source attribution in form of text, tabular or infobox proofs. We evaluate our dataset on state-of-the-art models in three different setting: oracle documents provided, closed, and open book setting. Our results and analysis suggest that TANQ is a complex task with many challenges ahead. 

\bibliography{anthology,custom,tacl2021}
\bibliographystyle{acl_natbib}

\appendix

\begin{figure*}[t]
    \centering
    \resizebox{1.0\textwidth}{!}{%
    \includegraphics{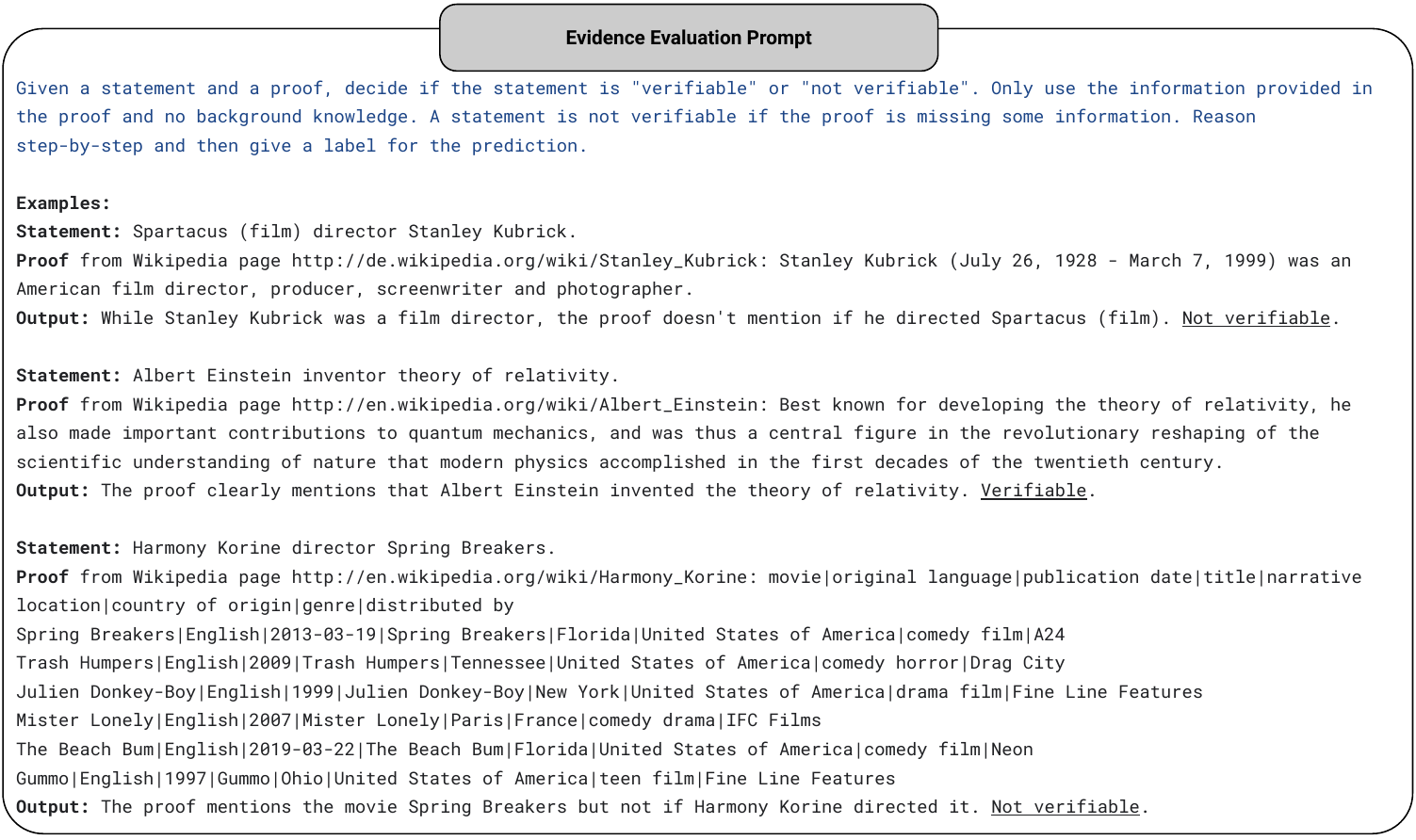}
    }
    \caption{Prompt used for evidence evaluation. We prompt a language model to evaluate the extracted evidence in  a natural language inference setting. The LLM labels the input statements as ``verifiable'' or ``not verifiable'' based on evidence provided in form of a sentence, table or infobox.}
    \label{fig:evidence_eval_prompt}
\end{figure*}

\begin{figure*}[t]
    \centering
    \resizebox{1.0\textwidth}{!}{%
    \includegraphics{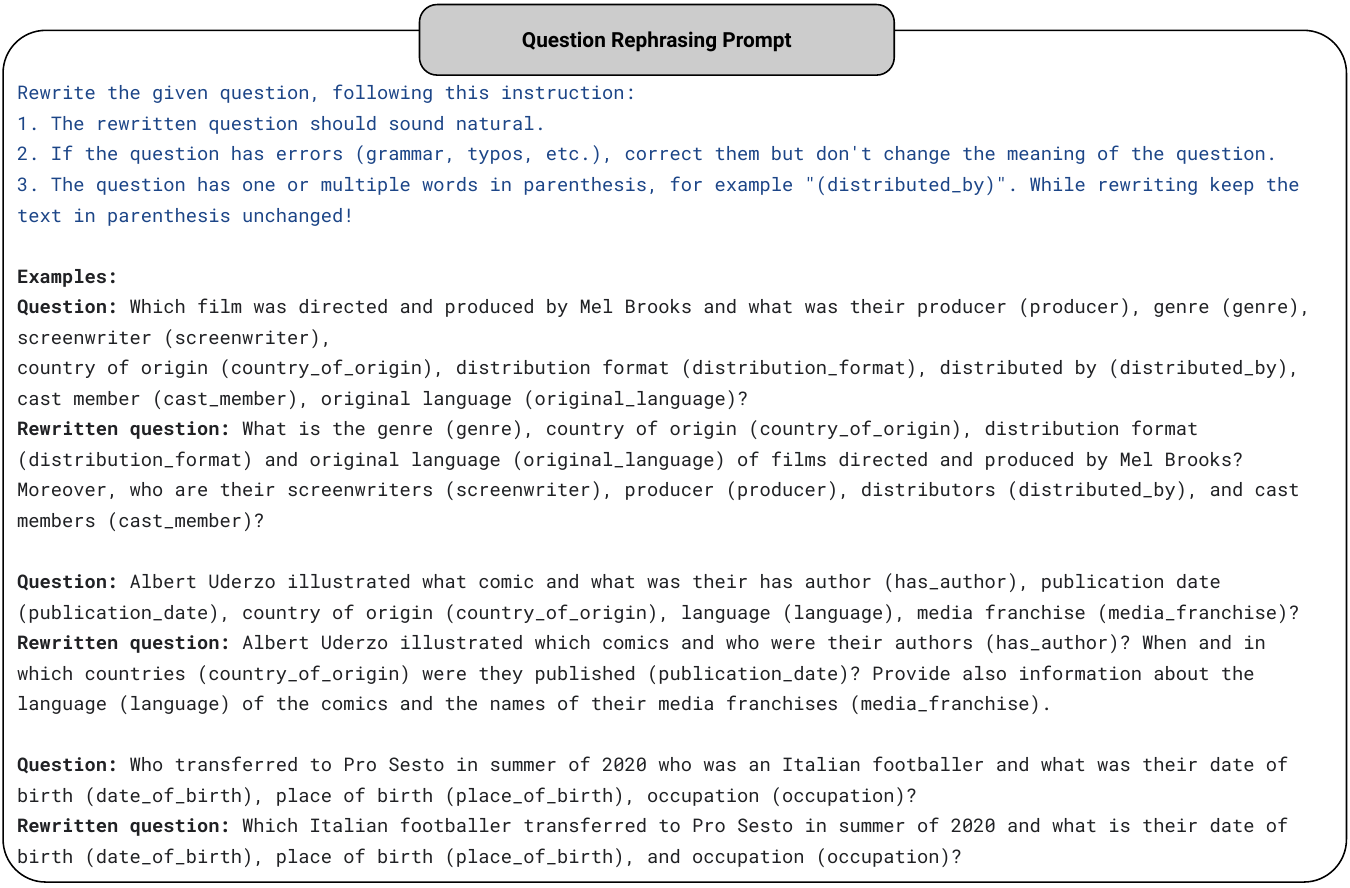}
    }
    \caption{Rephrasing questions increases the naturalness of template-based extension questions. To ensure the question meaning is preserved during rephrasing, we add structured annotations in parenthesis to questions with the name of each relation.}
    \label{fig:rephrasing_prompt}
\end{figure*}


\begin{table*}[h!]
\centering
\scalebox{0.7}{
\begin{tabular}{|l|l|l|l|l|l|l|}
\hline
Name & Occupation & Date of Death & Date of Birth & Citizenship & Place of Birth & Award Received \\ \hline
Mike Nichols & director, producer & November 19, 2014 & November 6, 1931 & United States & Berlin, Germany & Academy Award [...] \\ \hline
\textbf{Popeye} & Sailor &  & January 17, 1929 &  & & \\ \hline
\textbf{Munro} &  &  &  &  & & \\ \hline
\end{tabular}}
\caption{\label{tab:react_example_1}Example table for issues related to composition questions in open book baseline. While the question asks for directors, the second and third row contain movies.}
\end{table*}

\begin{table*}[h!]
\centering
\scalebox{0.6}{
\begin{tabular}{|l|l|l|l|l|l|l|}
\hline
& Director & Producer & Dir. of Photography & Composer & Genre & Publication \\ \hline
The Life and Death of Colonel Blimp & Michael Powell & E. Pressburger & Georges Périnal & Allan Gray & Drama, War & 1943 \\ \hline
A Canterbury Tale & Michael Powell & E. Pressburger & Erwin Hillier & Allan Gray & Drama, Romance & 1944 \\ \hline
A Matter of Life and Death & Jack Cardiff &  & Allan Gray & & Fantasy, War & 1946 \\ \hline
The Tales of Hoffmann & Jacques Offenbach &  &  &  &  &  \\ \hline
\end{tabular}}
\caption{\label{tab:react_example_3}Example table for issues related to missing attributes in answer tables. For the last row no relation other than director could be retrieved.}
\end{table*}

\begin{table*}
        \centering
        \scalebox{0.7}{
        \begin{tabular}{l c c c | c c c | c c c | c c c}
        \hline
        \bf Model	& \multicolumn{3}{c|}{\bf One Shot} & \multicolumn{3}{c|}{\bf Three Random Shots} & \multicolumn{3}{c|}{\bf Three Manual Shots} & \multicolumn{3}{c}{\bf Five Shots}\\
        \hline
         \bf  & \bf  P & \bf  R & \bf F1 & \bf  P & \bf  R & \bf F1 & \bf  P & \bf  R & \bf F1 & \bf  P & \bf  R & \bf F1  \\
        \hline
         \multicolumn{13}{c}{\bf Oracle setting} \\
        \hline
         Gemini Pro & 41.8 & 38.2 & 37.8 & 41.5 & 38.9 & 38.4 & 50.8 & 40.0 & 39.0 & 41.8 & 38.2 & 37.8 \\
         Gemini Flash & 54.7 & 56.8 & 50.8 & 66.8 & 59.0 & 60.7 & 67.1 & 57.7 & 59.4 & 54.7 & 58.2 & 53.3 \\
         GPT$4o$ & 55.4 & 44.9 & 46.6 & 54.3 & 44.9 & 45.9 & 39.9 & 34.8 & 35.4 & 48.8 & 40.1 & 40.7 \\
        \hline
         \multicolumn{13}{c}{\bf Closed book setting} \\
        \hline
         PaLM-$2$ & 46.4 & 42.9 & 43.2 & 52.9 & 46.4 & 47.6 & 52.3 & 45.8 & 47.0 & 52.1 & 45.8 & 47.0 \\
         Gemma & 28.0 & 18.2 & 19.2 & 38.8 & 26.3 & 27.5 & 40.9 & 30.6 & 31.8 & 38.5 & 33.1 & 34.6 \\
         Gemini Pro & 37.9 & 33.6 & 34.3 & 46.6 & 30.9 & 32.0 & 44.9 & 26.8 & 30.5 & 39.2 & 28.7 & 29.9 \\
         Gemini Flash & 21.3 & 25.9 & 18.6 & 36.8 & 28.5 & 30.7 & 33.5 & 30.2 & 30.1 & 29.9 & 28.6 & 23.9 \\
         GPT$4o$ & 37.2 & 23.6 & 26.3 & 37.9 & 22.4 & 24.6 & 66.1 & 15.7 & 16.2 & 38.1 & 18.9 & 20.4 \\
         \hline 
        \end{tabular}}
        \vspace{-0.5em}
        \caption{\small \label{tab:results_of_models_with_different_prompt_settings} 
        Evaluation results of models with different prompt settings. For most baselines the 3-shot setting yields the best results.}
\end{table*}

\begin{table*}
        \centering
        \scalebox{0.7}{
        \begin{tabular}{l c c c | c c c | c c c | c c c}
        \hline
        \bf Model	& \multicolumn{3}{c|}{\bf Empty Instruction} & \multicolumn{3}{c|}{\bf Detailed Instruction} & \multicolumn{3}{c|}{\bf Step-by-Step Instruction} & \multicolumn{3}{c}{\bf Simple Instruction}\\
        \hline
         \bf  & \bf  P & \bf  R & \bf F1 & \bf  P & \bf  R & \bf F1 & \bf  P & \bf  R & \bf F1 & \bf  P & \bf  R & \bf F1  \\
        \hline
         Oracle Setting & 61.8 & 55.4 & 56.4 & 66.8 & 60.2 & 61.8 & 67.2 & 60.3 & 61.6 & 66.8 & 59.0 & 60.7 \\
         Closed Book Seeting & 34.2 & 25.7 & 28.5 & 37.5 & 27.0 & 29.6 & 36.1 & 29.1 & 31.2 & 36.8 & 28.5 & 30.7 \\
        \hline
        \end{tabular}}
        \vspace{-0.5em}
        \caption{\small \label{tab:results_of_Gemini_Flash_with_different_prompt_instruction_styles} 
        Evaluation results of three-random-shots prompt with different instruction styles on Gemini Flash. We observe clear difference with/without instruction but almost the same result on different instruction styles. \Cref{fig:rephrasing_prompt} shows the text for the different instruction variants.}
\end{table*}

\begin{figure*}[t]
    \centering
    \resizebox{0.8\textwidth}{!}{%
    \includegraphics{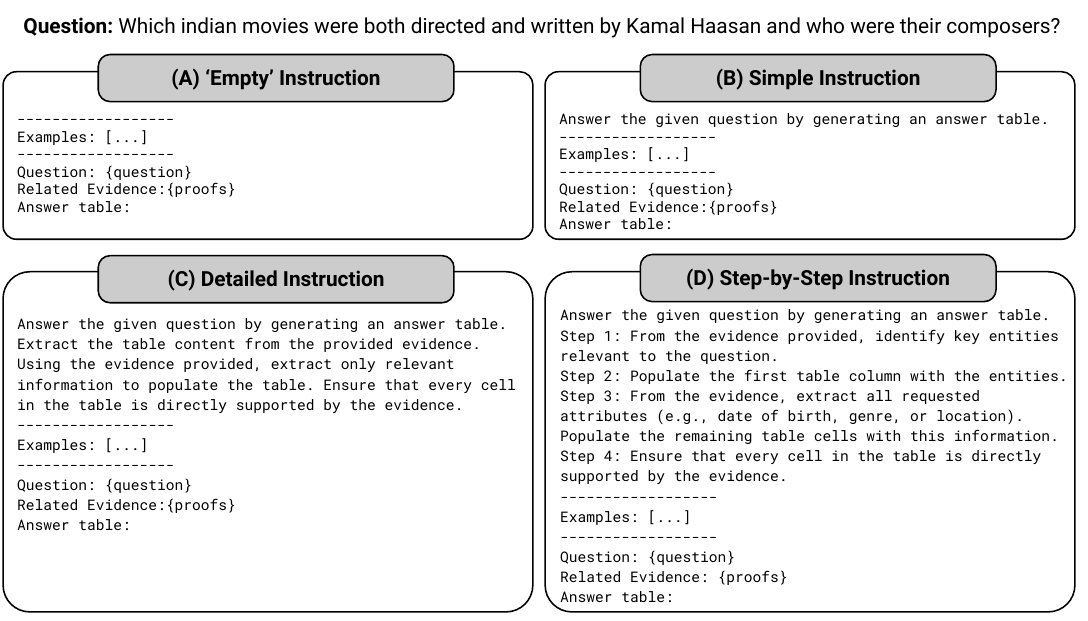}
    }
    \caption{Four instruction variants we evaluated: empty, simple, detailed, and step-by-step.}
    \label{fig:rephrasing_prompt}
\end{figure*}

\begin{figure*}[t]
    \centering
    \resizebox{0.7\textwidth}{!}{%
    \includegraphics{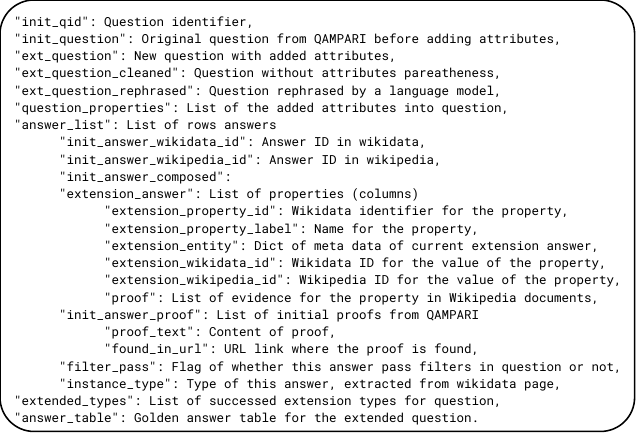}
    }
    \caption{Json schema description of a TANQ dataset entry outlining all components of a single dataset entry.}
    \label{fig:proto_defination}
\end{figure*}

\section{Modeling details}

For our experiments, we use the following GPT$4o$ model available over the OpenAI API: \texttt{gpt-4o-2024-08-06}.\footnote{\url{https://platform.openai.com/docs/models/gpt-4o}}
The PaLM-$2$ model we used for the pipeline and evaluation purposes was the model variant with $340$B parameters.

\end{document}